\documentclass[letterpaper]{article} 
\usepackage{aaai24}  
\usepackage{times}  
\usepackage{helvet}  
\usepackage{courier}  
\usepackage[hyphens]{url}  
\usepackage{graphicx} 
\usepackage{natbib}  
\usepackage{caption} 
\usepackage{algorithm}
\usepackage{algorithmic}

\usepackage{amsmath}
\usepackage{amssymb}
\usepackage{booktabs}
\usepackage{multirow}
\usepackage{colortbl}
\newcommand{\CheckmarkBold}{\textbf{\checkmark}}
\usepackage{arydshln}
\usepackage{xcolor}
\usepackage{tabularray}

\usepackage{newfloat}
\usepackage{listings}
\DeclareCaptionStyle{ruled}{labelfont=normalfont,labelsep=colon,strut=off} 
\floatstyle{ruled}
\newfloat{listing}{tb}{lst}{}
\floatname{listing}{Listing}
\title{Spectral Prompt Tuning: \\Unveiling Unseen Classes for Zero-Shot Semantic Segmentation}

\author{
    Wenhao Xu\textsuperscript{\rm 1,}\equalcontrib,
    Rongtao Xu\textsuperscript{\rm 2,3,}\equalcontrib,
    Changwei Wang\textsuperscript{\rm 2,3},
    Shibiao Xu\textsuperscript{\rm 1,}\thanks{
Shibiao Xu is the corresponding author. },
    \\
    Li Guo\textsuperscript{\rm 1},
    Man Zhang\textsuperscript{\rm 1},
    Xiaopeng Zhang\textsuperscript{\rm 2}
}
\affiliations{
    \textsuperscript{\rm 1}School of Artificial Intelligence, Beijing University of Posts and Telecommunications\\
     \textsuperscript{\rm 2}State Key Laboratory of Multimodal Artificial Intelligence Systems, Institute of Automation, Chinese Academy of Sciences\\
     \textsuperscript{\rm 3}School of Artificial Intelligence, University of Chinese Academy of Sciences\\
     shibiaoxu@bupt.edu.cn
     
}

\usepackage{bibentry}

\begin{document}

\maketitle

\begin{abstract}
Recently, CLIP has found practical utility in the domain of pixel-level zero-shot segmentation tasks. 
The present landscape features two-stage methodologies beset by issues such as intricate pipelines and elevated computational costs. While current one-stage approaches alleviate these concerns and incorporate Visual Prompt Training (VPT) to uphold CLIP's generalization capacity, they still fall short in fully harnessing CLIP's potential for pixel-level unseen class demarcation and precise pixel predictions.
To further stimulate CLIP's zero-shot dense prediction capability, we propose SPT-SEG, a one-stage approach that improves CLIP's adaptability from image to pixel.
Specifically, we initially introduce Spectral Prompt Tuning (SPT), incorporating spectral prompts into the CLIP visual encoder's shallow layers to capture structural intricacies of images, thereby enhancing comprehension of unseen classes.
Subsequently, we introduce the Spectral Guided Decoder (SGD), utilizing both high and low-frequency information to steer the network's spatial focus towards more prominent classification features, enabling precise pixel-level prediction outcomes.
Through extensive experiments on two public datasets, we demonstrate the superiority of our method over state-of-the-art approaches, performing well across all classes and particularly excelling in handling unseen classes.
\end{abstract}

\section{Introduction}
Semantic segmentation is one of the fundamental tasks in computer vision, aiming to predict the class for each pixel in an image~\cite{xu2023rssformer,xu2021dc,chen2021semi,9616392_Dong}. 
Despite the existence of numerous related works~\cite{lu2020video,What_Transferred_Dong_CVPR2020,xu2023dual,wang2023automatic}, the success of deep semantic segmentation models heavily relies on a large amount of annotated training images, which requires significant efforts. 
In recent years, interest has been growing in unsupervised or weakly supervised semantic segmentation methods, including semi-supervised~\cite{chen2021semi}, weakly supervised~\cite{xu2023scd,xu2023wave,wang2023treating}, few-shot~\cite{xie2021scale}, and zero-shot semantic segmentation~\cite{bucher2019zero,pastore2021closer,xian2019semantic}. 
Among them, zero-shot semantic segmentation tasks are particularly challenging and appealing, as they require generating accurate semantic segmentation results with only the semantic descriptions of the classes given.

\begin{figure}[!t]
\centering
 \includegraphics[width=8cm,height=5cm]{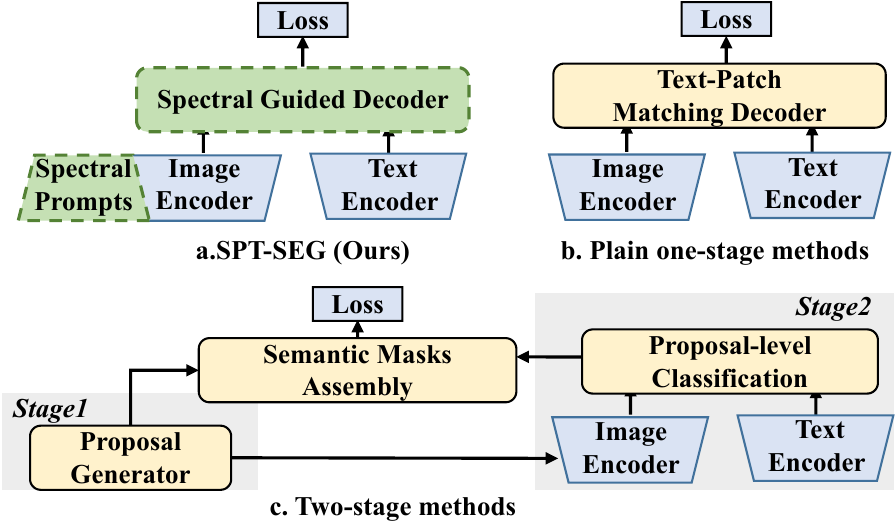}
\caption{
(a) Our SPT-SEG method demonstrates outstanding performance across all classes.
(b) While yielding favorable results within the seen classes, it exhibits relatively poorer performance in the unseen classes.
(c) Its performance is unsatisfactory across all classes.}
\label{fig:intro}
\end{figure}

To incorporate zero-shot capability into visual systems, researchers have proposed large-scale vision-and-language pretraining models, such as CLIP~\cite{clip} and ALIGN~\cite{jia2021scaling}. 
Specifically, CLIP encodes semantic concepts into model parameters by contrastive training on a massive collection of image-text pairs, forming a zero-shot knowledge base for downstream tasks. 
However, contrastive pretraining mainly focuses on capturing image-level concepts. In CLIP, the training texts primarily describe the global context of images, and the encoded image and text embeddings are used together to compute contrastive losses. Consequently, CLIP is more suitable for image-level classification tasks~\cite{zhou2022learning,zhou2022conditional,lu2022prompt,zhang2022tip}.
The pretrained visual-language model CLIP~\cite{clip} has recently found applications in various dense prediction tasks, including semantic segmentation~\cite{pak2021seg}, referring segmentation~\cite{wang2022cris}, and object detection~\cite{pour2022zero}. 
In the zero shot semantic segmentation task, approaches like zsseg~\cite{xu2021simple} and Zegformer~\cite{ding2022decoupling} adopt a similar strategy that requires two-stage processing: first generating region proposals and then feeding the cropped regions into CLIP for zero-shot classification. 
However, this strategy involves encoding images twice as FI~\ref{fig:intro}(c), once for proposal generation and another for CLIP encoding of each proposal. This design introduces additional computational overhead and fails to fully leverage the knowledge in the CLIP encoder to guide the proposal generation stage. 
To streamline the process, ZegCLip~\cite{zegclip} introduces a one-stage approach by incorporating visual prompt tuning into CLIP, then extending CLIP's zero-shot capabilities from image-level to pixel-level.

The inclusion of Visual Prompt Tuning (VPT) in CLIP significantly enhances its downstream task generalization with few learnable parameters. However, since the original CLIP's training primarily revolves around image-level contrastive learning, its features tend to emphasize only the most discriminative parts of objects. 
Even with the introduction of VPT, the observed phenomenon persists even during pre-training with image-level contrastive loss. Consequently, this phenomenon leads to incomplete and biased segmentation in dense prediction tasks.

Based on the aforementioned observations, we believe that further enhancing the image-to-pixel adaptability of CLIP~\cite{clip} would contribute to improved zero-shot segmentation performance.
Therefore, we propose an innovative one-stage method called SPT-SEG, as shown in Fig.~\ref{fig:intro}(b). 
SPT-SEG differs from plain one-stage methods, as depicted in Fig.\ref{fig:intro}(a). 
In our approach, we integrate spectral cues into the shallow layers of the CLIP visual encoder, which provides additional structural information that enhances the model's comprehension of various object components. 
We also utilize high-frequency and low-frequency information to guide the alignment of text and pixels, directing the network's spatial focus towards more salient classification features. 
The synergy of these two designs enhances the model's semantic understanding and reasoning capabilities, effectively addressing the issues of inadequate pixel generalization and incomplete segmentation present in the current CLIP-based zero-shot semantic segmentation methods.

In summary, our contributions are listed as follows:
\begin{itemize}
\item We introduce \textbf{Spectral Prompt Tuning (SPT)}, which builds upon VPT by incorporating a small set of learnable spectral parameters. These parameters are integrated into the shallow layers of the CLIP visual encoder to introduce spectral information.

\item We propose the \textbf{Spectral Guided Decoder (SGD)} layer, which is a novel component that utilizes high-frequency and low-frequency information to guide the matching process between textual and pixel representations. 

\item We comprehensively assess our method on two public datasets, and the results clearly show that our approach significantly surpasses state-of-the-art methods.
\end{itemize}

\begin{figure*}[!ht]
    \centering
   \includegraphics[width=17.5cm,height=8.5cm]{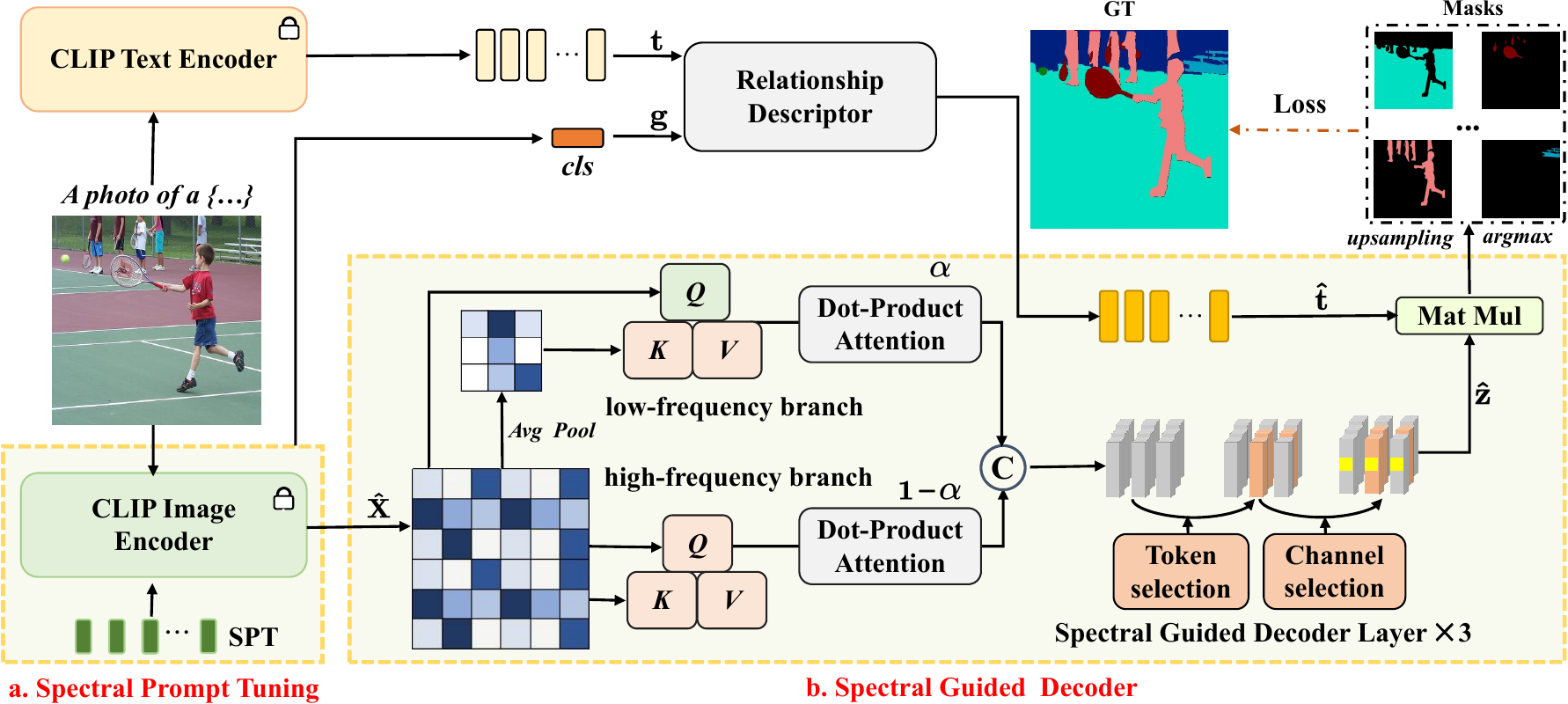}
    \caption{Overview of our proposed SPT-SEG. The main contribution of our work lies in two simple but effective designs (Red marks a,b in the figure): (a) Spectral prompt tuning which adds learnable spectral prompts to the first two layers of the CLIP's visual encoder; (b) Spectral guided decoder which utilizes high- and low-frequency feature information to guide the text to match with pixels, and decodes the predicted results. }.
   \label{fig:spt} 
\end{figure*}

\section{Related Work}
\noindent
\textbf{Vision-Language Model.}
Extensive research has been conducted on Visual-Language Models (VLM)\cite{hong2021vln, huang2021seeing, kamath2021mdetr, kim2021vilt}, showcasing significant advancements in downstream vision tasks, especially in settings with unannotated or restricted data. 
These tasks encompass diverse areas such as image retrieval\cite{liu2021image}, dense prediction~\cite{rao2022denseclip}, visual referring expression~\cite{wang2022cris}, and visual question answering~\cite{jiang2022finetuning}. 
CLIP ~\cite{clip} is widely recognized as one of the most popular vision-language models.
It is pretrained using contrastive learning on a massive dataset of 400 million text-image pairs. 
ALIGN~\cite{align} utilized an even larger dataset, comprising 1.8 billion pairs, for pre-training its model. However, this larger dataset also introduced a significant amount of noise.
In more recent works, CoCa ~\cite{coca} and Beit-V3 ~\cite{beitv3} have further emphasized the superior performance of VLM pre-trained features.

\noindent
\textbf{Prompt Tuning.}
The concept of prompts originated from natural language processing and is mainly used in VLM to enhance its understanding of downstream specific tasks. By providing prompts, we can avoid massive parameter learning for VLM and instead use it as a fixed knowledge base, focusing only on task-relevant information. These prompts can be manually created for downstream tasks or automatically learned during fine-tuning. 
Full fine-tuning and linear probe~\cite{gao2021clip} are two typical methods for adapting the VLM (i.e. CLIP) to downstream tasks.
Full fine-tuning leads to a reduced VL representation of previously learned, while linear probe limits the zero-shot capability of CLIP. 
Inspired by the prompt learning in NLP, many works propose to adapt VLM by adding learnable tokens during end-to-end training.
CoOp~\cite{zhou2022learning} introduced continuous prompt learning, where a set of continuous vectors are optimized end-to-end with down-stream supervision . Additionally, learnable prompts are applied by CoOp on the text encoder of CLIP to replace sub-optimal hand-crafted templates. 
Co-CoOp~\cite{zhou2022conditional} highlights the poor performance of CoOp on novel classes and addresses the generalization problem by explicitly conditioning the prompts on image instances.
Recently, prompting~\cite{jia2022visual,sandler2022fine} has been adapted to vision tasks. \cite{sandler2022fine} proposes memory tokens which is a set of learnable embedding vectors for each transformer layer. 
VPT~\cite{jia2022visual} proposes similar ideas and investigates the generality and feasibility of visual prompting via extensive experiments spanning multiple kinds of recognition tasks across multiple domains and backbone architectures. 
Our research further extends the paradigm of visual prompt learning by introducing spectral prompt, addressing the limitations of previous visual prompt learning methods in fully leveraging the structural information of images and their limited adaptability to pixel-level tasks.

\noindent
\textbf{Zero-shot Semantic Segmentation.}
It remains a challenging task to achieve zero-shot semantic segmentation due to the presence of an imbalance problem in seen classes.
Previous studies such as SPNet~\cite{xian2019semantic}, ZS3~\cite{bucher2019zero}, CaGNet~\cite{gu2020context} and STRICT~\cite{pastore2021closer} adopt strategies to improve the generalization ability of semantic mappings from visible to invisible classes.
Since the popular pre-trained visual language model CLIP has shown powerful zero-shot classification capabilities, it has recently been applied to zero-shot semantic segmentation as well.
Zegformer~\cite{ding2022decoupling} and zsseg~\cite{xu2021simple} developed an extensive proposal generator and used CLIP to classify each region and then integrate the predictions.
Previous studies, such as SPNet~\cite{xian2019semantic}, ZS3~\cite{bucher2019zero}, CaGNet~\cite{gu2020context}, SIGN~\cite{cheng2021sign}, Joint~\cite{baek2021exploiting}, and STRICT~\cite{pastore2021closer}, adopt the approach of improving the generalization capability of semantic mapping from the classes that have been encountered to unseen ones.
Recently, a two-stage paradigm~\cite{ding2022decoupling,xu2021simple} has been proposed to explore the use of CLIP for zero-shot segmentation.
They leveraged the CLIP model to classify individual regions following a comprehensive proposal generator and then integrate the resulting predictions.
Although effective, this design requires two image encoding processes, resulting in expensive computational costs. 
In order to simplify the pipeline of the two stages, ZegCLIP~\cite{zegclip} proposed a one-stage method that transfers CLIP's powerful generalization ability from images to pixel-level classification.
In this work, we use a one-stage method and achieve outstanding zero-shot segmentation performance through two effective designs.

\section{Method}
\subsection{Problem Definition}
We adopt the generalized zero-shot semantic segmentation (GZLSS) method~\cite{xian2019semantic}, which requires to segment both seen classes ${C}^{s}$ and unseen classes ${C}^{u}$ after only training on a dataset with pixel-annotations of seen part.
During training, the model generates per-pixel classification results based on the semantic descriptions of all visible classes. During testing, the model is evaluated on both seen and unseen classes. It is important to note that $\mathcal{C}^{s} \cap \mathcal{C}^{u}=\oslash$ and that the label of $\mathcal{C}^{u}$ is not available during training.

\subsection{SPT-SEG}
The architecture of SPT-SEG is illustrated in Fig.~\ref{fig:spt}. 
The basic one-stage methodology comprises four key components: the CLIP encoder that incorporates the text and visual encoders, the relationship descriptor between the cls token and the text embeding, a decoder, and a loss function.
Our enhancements focus on two pivotal components: (1) Introducing an innovative \textbf{Spectral Prompt Tuning} approach within the visual encoder, aimed at extracting structural insights to bolster CLIP's adaptability to dense prediction tasks, (2) Integrating a \textbf{Spectral Guided Decode Layer} into the decoder, which adeptly captures high and low-frequency features specific to the task.

\begin{figure}[!t]
\centering
\includegraphics[width=6.9cm,height=8.5cm]{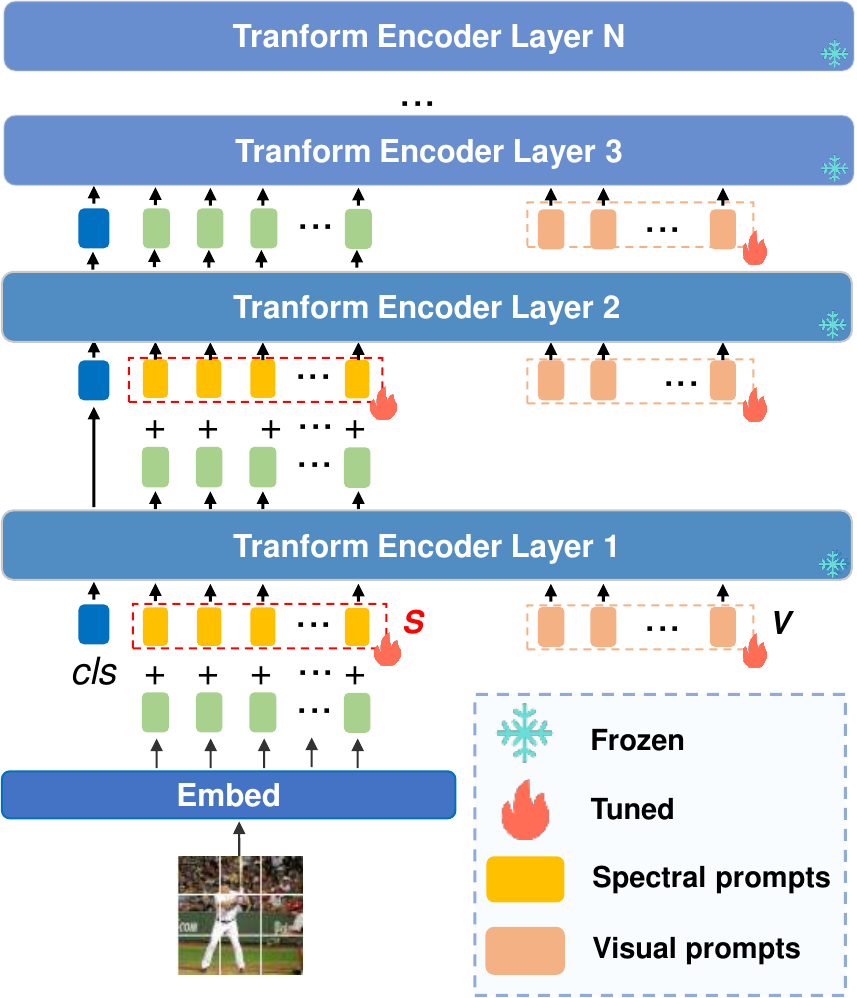}
\caption{Overview of our proposed Spectral-Prompt Tuning. During training on downstream tasks, only the parameters of prompts and the linear head are updated while the whole Transformer encoder is frozen.}
\label{fig:spt_detail} 
\end{figure}
\subsubsection{Spectral Prompt Tuning}\label{sec:spt}
Prompt tuning is a recently proposed fine-tuning technique that offers a valuable approach to adapt pre-trained transformer models to target domains~\cite{xing2022class}.
However, fine-tuning zero-shot segmentation models solely on a limited set of visible classes often leads to overfitting. This occurs because the optimization process focuses solely on visible classes, disregarding knowledge relevant to visual concepts that cannot be obtained from the training set. To address this issue, Visual Prompt Tuning (VPT)~\cite{jia2022visual} has emerged as a potential solution. VPT introduces a small number of task-specific learnable parameters in the input space while keeping the backbone frozen during downstream training. While VPT has shown promising results in certain cases, it does not fully leverage the intrinsic properties and structural characteristics of images, which may not be fully manifested in the spatial domain, thereby limiting its effectiveness in handling structure-aware tasks.

To address this limitation, we propose the Spectral Prompt Tuning (SPT) method, as shown in Fig.~\ref{fig:spt_detail}. SPT extends the concept of VPT by incorporating prompt parameters learned from a spectral perspective. 

In contrast to VPT's exclusive reliance on visual prompts for fine-tuning, SPT capitalizes on frequency domain features to offer supplementary understanding of intricate attributes and structural characteristics. 
The features learned by SPT in the spectrum allow it to better capture and distinguish subtle visual features of different classes, even for those classes that do not have direct examples in the training data. In this way, when the model encounters images of completely new classes, it can extract common information about these classes from the spectrum features, enabling more accurate segmentation. This ability can alleviate the "partial" or "ambiguous" segmentation issues that occur in zero-shot scenarios, thus ensuring a more precise capture of unknown classes.

The input embeddings from the $l$-th layer of the image encoder in the CLIP model are denoted as $\left\{\mathbf{g}^{l}, \mathbf{h}_{1}^{l}, \mathbf{h}_{2}^{l}, \cdots, \mathbf{h}_{N}^{l}\right\}$.
Here, $\mathbf{g}^{l}$ represents the embedding for the [cls] token, and $\mathbf{H}^{l}=\left\{\mathbf{h}_{1}^{l}, \mathbf{h}_{2}^{l}, \cdots, \mathbf{h}_{N}^{l}\right\}$ corresponds to the embeddings of image patches.
In the context of SPT, the CLIP image encoder's token sequences are extended with learnable tokens $\mathbf{V}^{l}=\left\{\mathbf{v}_{1}^{l}, \mathbf{v}_{2}^{l}, \cdots, \mathbf{v}_{M}^{l}\right\}$ in each layer. 
Furthermore, learnable spectral prompts $\mathbf{S}^{l}=\left\{\mathbf{s}_{1}^{l}, \mathbf{s}_{2}^{l}, \cdots, \mathbf{s}_{N}^{l}\right\}$ are added in the first two layers.
These additions enhance the model's ability to process image features at multiple levels of abstraction.

$\mathbf{S}^{l}$ is calculated from $\mathbf{H}^{l}$ and $\mathbf{g}^{l}$, and a set of learnable filter parameters $\mathbf{w}_{f}$, the process can be expressed as:

\begin{align}
    \mathbf{S}^{l}=\operatorname{\mathcal{F}^{-1}}(\operatorname{\mathcal{F}}(\mathbf{H}^{l}\odot \mathbf{g}^{l} ) \odot \mathbf{w}_{f}),
\end{align}
where $\mathcal{F}$ is the 2D fast fourier transform (FFT) and $\mathcal{F}^{-1}$ is the inverse FFT (IFFT).
Then, when $l \le 2$ the layer processes the input token as: 

\begin{align}
   \left[\mathbf{g}^{l},{}_{-}, \mathbf{H}^{l}\right]= \operatorname{Layer}^{l} \left(\left[\mathbf{g}^{l-1}, \mathbf{V}^{l-1}, \mathbf{H}^{l-1}+\mathbf{S}^{l-1} \right] \right) ),
\end{align}
when $l > 2$ the transform layer processes the input token as:

\begin{align}
     \left[\mathbf{g}^{l},{ }_{-}, \mathbf{H}^{l}\right]= \operatorname{Layer}^{l} \left(\left[\mathbf{g}^{l-1}, \mathbf{V}^{l-1}, \mathbf{H}^{l-1}\right]\right) ).
\end{align}

\subsubsection{Spectral Guided Decode Layer}\label{sec:sgd}
In practical semantic segmentation applications, high-quality segmentation results are crucial for the success of the task.
 Recent work~\cite{patro2023spectformer} combined spectral layers with multi-head attention in a transformer architecture to capture relevant features in initial layers.
LiTv2 ~\cite{pan2022hilo}introduced a novel attention mechanism that separately processes high and low-frequency components in attention layers, capturing local and global relationships effectively in classification and segmentation tasks. 
Drawing inspiration from these insights, we propose an innovative decoding method as shown Fig.~\ref{fig:spt}(b) by introducing frequency domain features during the decoding stage, which significantly enhances the performance of image segmentation. 
Firstly, the frequency domain-guided decoder can balance the attention on small details and global structure, enabling the model to focus on both local and overall features simultaneously. 
Secondly, guided by frequency domain features, the decoder can capture object boundaries and textures more accurately, thereby improving the precision of the segmentation results. Most importantly, this decoder exhibits stronger generalization ability on unseen classes, which is crucial for unknown situations in real-world applications.
The design comprises the following steps:

(1) The high-frequency branch captures fine-grained local dependencies through local window self-attention., while the low-frequency branch applies average pooling to each window, obtaining low-frequency signals that capture the global dependencies of the input. This high and low-frequency capturing is built on the multi-head self-attention (MSA) mechanism, which allowsfor capturing  distant relations labeled at different locations in the input sequence $\mathbf{X} \in \mathbb{R}^{N \times D}$. Here, $N$ is the length of the input sequence, and $D$ represents the hidden dimension.
To achieve this, we divide the $N_{h}$ heads in MSA into two groups with a split ratio $\alpha$. 
Specifically, $\alpha N_{h}$ heads are used for the high-frequency branch, and the remaining $(1-\alpha)N_{h}$ heads are utilized for the low-frequency branch. 
The high-frequency branch computes the output by linearly projecting the outputs of the $\alpha$ self-attention heads and then concatenating them as follows:
The high-frequency branch performs a simple non-overlapping-window ($3 \times 3$) partitioning of the inputs $X$, and then computes the outputs by $\alpha$-sizing and concatenating them as follows:

\begin{equation}
\mathrm{MSA_{\alpha}}(\hat{\mathbf{X}}) = \underset{h\in [\alpha N_{h}]}{\mathrm{Concat}}[\mathrm{SA}_h(\hat{\mathbf{X}})],
\end{equation}
where $\mathrm{SA}_h(\hat{\mathbf{X}})$ denotes the output of the $h$-th self-attention head, and note that $\hat{\mathbf{X}} $ denotes the input with the non-overlapping window already divided.
Meanwhile, the low-frequency branch utilizes average pooling to extract low-frequency signals within each window, and its computation process can be expressed as:

\begin{equation}
\mathrm{MSA_{1-\alpha}}(\hat{\mathbf{X}}) = \underset{h\in [(1-\alpha)N_{h}]}{\mathrm{Concat}}[\mathrm{SA}_h(\mathrm{AvgPool}(\hat{\mathbf{X}}))],
\end{equation}
Finally, the overall output is obtained by concatenating the outputs from each branch as follows:

\begin{equation}
\mathbf{z} = [\mathrm{MSA_{\alpha}}(\hat{\mathbf{X}}); \mathrm{MSA_{1-\alpha}}(\hat{\mathbf{X}})],
\end{equation}
where $[\cdot]$ denotes the concatenation operation.

(2) we emphasize task-relevant tokens and channels through frequency domain feature extraction to select specific characteristics. We perform frequency domain feature extraction on $\mathbf{z} \in \mathbb{R}^{N \times D}$ to identify task-related markers and channels. The output is obtained using the following operation:

\begin{align}
\mathbf{\hat{z}} = P \cdot \text{sim}(\mathbf{z}, \xi) \cdot \mathbf{z},
\end{align}
where $\xi \in \mathbb{R}^d$ and $P \in \mathbb{R}^{d \times d}$ are task-specific parameters, and $\text{sim}(\cdot, \cdot)$ represents the cosine similarity ranging between $[0, 1]$.
The resulting $\hat{\mathbf{z}}$ can be represented as $[\hat{\mathbf{z}}_{1}, \hat{\mathbf{z}}_{2}, ..., \hat{\mathbf{z}}_{N}] \in \mathbb{R}^{N \times D}$, where $\hat{\mathbf{z}}_{j}$ denotes the embedding for the j$^{th}$ patch class. The matrix $\mathbf{t} = [\mathbf{t}^{1}, \mathbf{t}^{2}, ..., \mathbf{t}^{C}] \in \mathbb{R}^{C \times D}$ represents $C$ classes, with $d$ as the feature dimension of the CLIP model. Here, $\mathbf{t}^{i}$ denotes the representation of the $i$-th class, and [cls] corresponds to the global feature represented as $\mathbf{g} \in \mathbb{R}^{N \times D}$.
The relationship descriptor can be represented as:

\begin{align}
    \mathbf{\hat{t}}=\phi(\mathbf{[t \cdot g; t]}),
\end{align}
where $\phi(\cdot)$ projects $\mathbf{[t \cdot g; t]}$ to the same dimension as $\hat{\mathbf{z}}$.

Semantic masks are calculated using matrix product:

\begin{equation}
\label{eq:decoder}
\mathbf{Masks} =   \mathbf{\hat{t}} \cdot \hat{\mathbf{z}}^{T}   \in \mathbb{R}^{C \times N},
\end{equation}
The final segmentation results are obtained by applying the $Argmax$ operation along the class dimension of $\mathbf{Masks}$.

\subsubsection{Loss Function}
We employ a combination of the focal loss~\cite{lin2017focal}, and the structural similarity (SSIM) loss~\cite{wang2003multiscale}. 
The total loss $\mathcal{L}$ is a linear combination of the focal loss and SSIM loss, with coefficients $\alpha$ and $\beta$ to balance their contributions:

\begin{equation}
\mathcal{L} = \gamma \cdot \mathcal{L}_{\mathtt{focal}} + \sigma \cdot \mathcal{L}_{\mathtt{ssim}},
\end{equation}
The coefficients ${\gamma, \sigma}$ are used to control the relative importance of the focal loss and SSIM loss in the overall loss function. 

\section{Experiments}
\subsection{Datasets}
We conducted extensive experiments on two benchmark datasets to evaluate the effectiveness of our proposed method: PASCAL VOC 2012 (20), COCO-Stuff 164K.
Here are the details of each dataset:

\begin{enumerate}
\item \noindent \textbf{PASCAL VOC 2012}: This dataset consists of 10,582 augmented images for training and 1,449 for validation. We focus on 15 seen classes, ignoring the "background" class, and 5 unseen classes.
\item \noindent \textbf{COCO-Stuff 164K}: It is a large-scale dataset with 118,287 training images and 5,000 testing images, covering 171 classes. Among them, 156 classes are seen, and 15 classes are unseen.

\end{enumerate}

\subsection{Evaluation Metrics}
As in previous studies, we assess the performance using pixel-wise classification accuracy ($pAcc$) and the mean intersection over union ($mIoU$) for both seen and unseen classes, referred to as $mIoU(S)$ and $mIoU(U)$, respectively. Additionally, we calculate the harmonic mean IoU ($hIoU$)  between the seen and unseen classes as in ZegCLIP~\cite{zegclip}, which is formulated as:

\begin{equation}
hIoU=\frac{2*mIoU(S)*mIoU(U)}{mIoU(S)+mIoU(U)}.
\end{equation}

\begin{table*}
\centering
\begin{tabular}{c|cccc|cccc} 
\toprule
\multirow{2}{*}{\textbf{Methods}}                           & \multicolumn{4}{c}{\textbf{PASCAL VOC 2012}}                                                                                                                                                                                                                                                                                                                                                                                                  & \multicolumn{4}{c}{\textbf{COCO-Stuff 164K}}                                                                                                                                                                                                                                                                                                                                                                                                   \\ 
\cline{2-9}
                                                            & \textbf{pAcc}                                                                                      & \textbf{mIoU(S)}                                                                                   & \textbf{mIoU(U)}                                                                                                               & \textbf{hIoU}                                                                                      & \textbf{pAcc}                                                                                      & \textbf{mIoU(S)}                                                                                   & \textbf{mIoU(U)}                                                                                                               & \textbf{hIoU}                                                                                       \\ 
\hline\hline
SPNet$_{CVPR'19}$                                             & /                                                                                                  & 78.0                                                                                               & 15.6                                                                                                                           & 26.1                                                                                               & /                                                                                                  & 35.2                                                                                               & 8.7                                                                                                                            & 14.0                                                                                                \\
ZS3$_{NeurIPS'19}$                                            & /                                                                                                  & 77.3                                                                                               & 17.7                                                                                                                           & 28.7                                                                                               & /                                                                                                  & 34.7                                                                                               & 9.5                                                                                                                            & 15.0                                                                                                \\
CaGNet$_{ACMMM'20}$                                           & 80.7                                                                                               & 78.4                                                                                               & 26.6                                                                                                                           & 39.7                                                                                               & 56.6                                                                                               & 33.5                                                                                               & 12.2                                                                                                                           & 18.2                                                                                                \\
SIGN$_{ICCV'21}$                                              & /                                                                                                  & 75.4                                                                                               & 28.9                                                                                                                           & 41.7                                                                                               & /                                                                                                  & 32.3                                                                                               & 15.5                                                                                                                           & 20.9                                                                                                \\
Joint$_{ICCV'21}$                                             & /                                                                                                  & 77.7                                                                                               & 32.5                                                                                                                           & 45.9                                                                                               & /                                                                                                  & /                                                                                                  & /                                                                                                                              & /                                                                                                   \\
ZegFormer$_{CVPR'22}$                                         & /                                                                                                  & 86.4                                                                                               & 63.6                                                                                                                           & 73.3                                                                                               & /                                                                                                  & 36.6                                                                                               & 33.2                                                                                                                           & 34.8                                                                                                \\
ZSSeg$_{arXiv'21}$                                            & 90.0                                                                                               & 83.5                                                                                               & 72.5                                                                                                                           & 77.5                                                                                               & 60.3                                                                                               & 39.3                                                                                               & 36.3                                                                                                                           & 37.8                                                                                                \\
ZegCLIP$_{CVPR'23}$                                           & 94.6                                                                                               & 91.9                                                                                               & 77.8                                                                                                                           & 84.3                                                                                               & 62.0                                                                                               & 40.2                                                                                               & 41.4                                                                                                                           & 40.8                                                                                                \\
\rowcolor[rgb]{0.922,0.918,0.918} \textbf{SPT-SEG (Ours)}   & \begin{tabular}[c]{@{}>{\cellcolor[rgb]{0.922,0.918,0.918}}c@{}}\textbf{96.7}\\(+2.1)\end{tabular} & \begin{tabular}[c]{@{}>{\cellcolor[rgb]{0.922,0.918,0.918}}c@{}}\textbf{92.9}\\(+1.0)\end{tabular} & \begin{tabular}[c]{@{}>{\cellcolor[rgb]{0.922,0.918,0.918}}c@{}}\textbf{87.4}\\({\textbf{+9.6}})\end{tabular} & \begin{tabular}[c]{@{}>{\cellcolor[rgb]{0.922,0.918,0.918}}c@{}}\textbf{90.1}\\(+5.8)\end{tabular} & \begin{tabular}[c]{@{}>{\cellcolor[rgb]{0.922,0.918,0.918}}c@{}}\textbf{62.9}\\(+0.9)\end{tabular} & \begin{tabular}[c]{@{}>{\cellcolor[rgb]{0.922,0.918,0.918}}c@{}}\textbf{40.6}\\(+0.4)\end{tabular} & \begin{tabular}[c]{@{}>{\cellcolor[rgb]{0.922,0.918,0.918}}c@{}}\textbf{43.8}\\({\textbf{+2.4}})\end{tabular} & \begin{tabular}[c]{@{}>{\cellcolor[rgb]{0.922,0.918,0.918}}c@{}}\textbf{42.1}\\(+1.3)\end{tabular}  \\ 
\hline
ZegCLIP *$_{CVPR'23}$                                         & 96.3                                                                                               & 92.4                                                                                               & 90.9                                                                                                                           & 91.6                                                                                               & 69.9                                                                                               & 40.7                                                                                               & 63.2                                                                                                                           & 49.6                                                                                                \\
\rowcolor[rgb]{0.922,0.918,0.918} \textbf{SPT-SEG * (Ours)} & \textbf{97.6}                                                                                      & \textbf{93.6}                                                                                      & \textbf{92.9}                                                                                                                  & \textbf{93.2}                                                                                      & \textbf{72.5}                                                                                      & \textbf{41.6}                                                                                      & \textbf{66.0}                                                                                                                  & \textbf{51.0}                                                                                       \\
\hline
\end{tabular}
\caption{Comparison with state-of-the-art methods on the PASCAL VOC 2012 and COCO-Stuff 164K datasets. The asterisk (*) denotes training involving all classes. The best results are highlighted in bold.}
\label{tab: results}
\end{table*}

\subsection{Implementation Details}
Our proposed method is implemented using the MMSegmentation open-source toolbox\cite{mmseg2020} with PyTorch 1.10.1. All experiments were conducted on two H800 GPUs using the pre-trained CLIP ViT-B/16 model. 
The batch size was set to 16, and the images were resized to a resolution of $512 \times 512$.
We performed a total of 20,000 training iterations on the PASCAL VOC 2012 dataset,  and 96,000 iterations on the COCO-Stuff 164K dataset. 
Based on previous research works \cite{gu2020context, xu2021simple, ding2022decoupling, zhou2022extract}, we have set up the unseen classes.
The optimizer used was AdamW, and we followed the default training schedule provided by the MMSeg toolbox.
In SPT-SEG, it should be noted that the model learns multiple prompts exclusively from seen classes during training. 
The optimizer used was AdamW, and we followed the default training schedule provided by the MMSeg toolbox.
\subsection{Comparison with State-of-the-Art Methods}
\begin{figure*}[!ht]
    \centering
    \includegraphics[width=17cm,height=9cm]{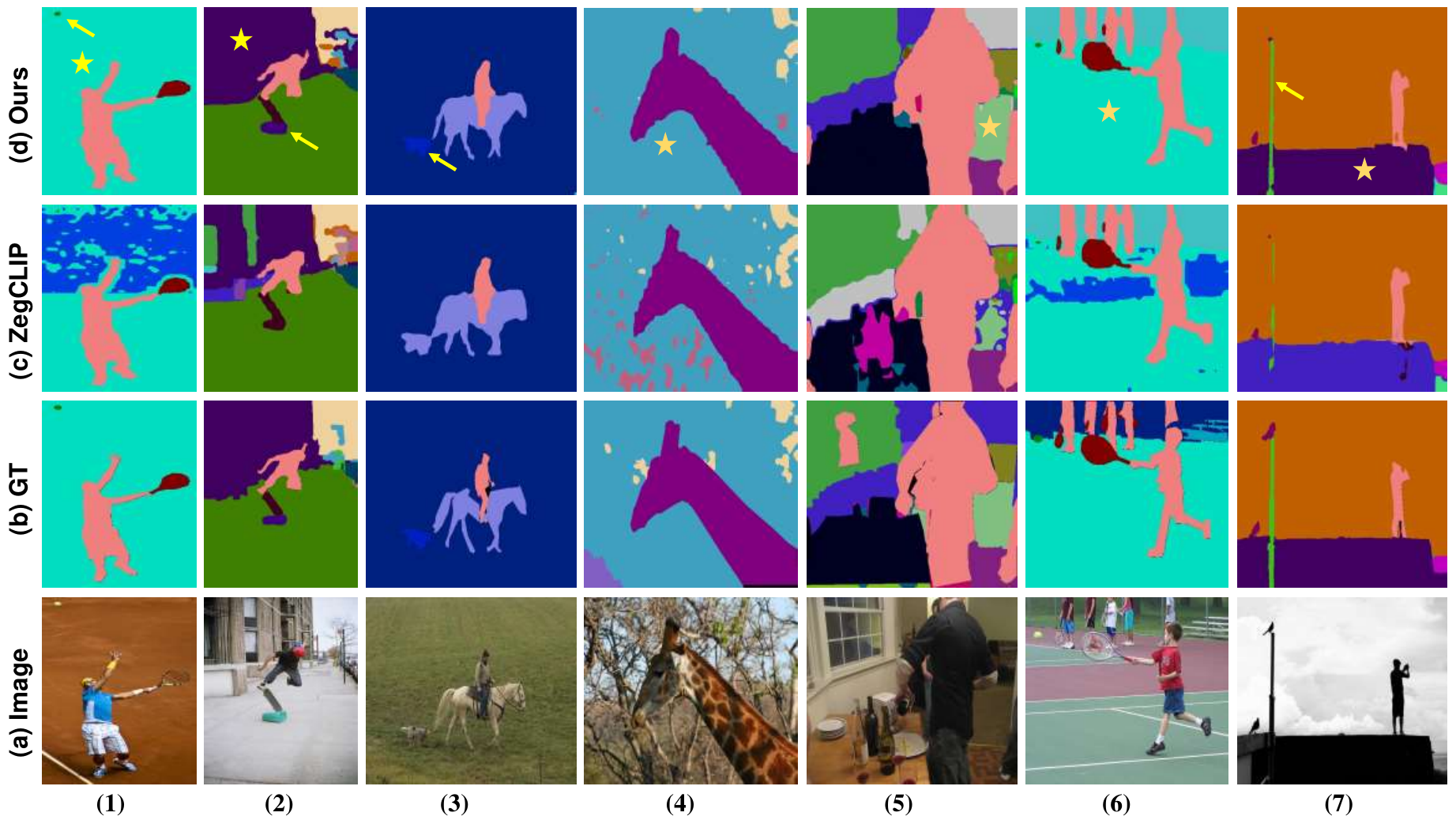}
    \caption{Qualitative results on COCO-Stuff 164K. (a) are the original testing images; (b) are the ground truths of each image.(c) represent the performance of ZegCLIP; (d) are the visualization results of our proposed SPT-SEG. Note that we have highlighted prominent regions using yellow arrows and marked other significant areas with yellow stars for emphasis.
}
   \label{fig:vis} 
\end{figure*}

To showcase the effectiveness of our method, we present the evaluation results in comparison with previous state-of-the-art approaches, as shown in Tab.~\ref{tab: results}. 
Additionally, we include the results of fully supervised learning as an upper bound to demonstrate the performance gap between fully supervised segmentation and zero-shot segmentation on unseen classes.
We provide qualitative results on the COCO-Stuff 164K dataset, depicted in Fig.~\ref{fig:vis}.
Our proposed method exhibits significant performance improvements, particularly for unseen classes, surpassing previous approaches, as depicted in Tab.~\ref{tab: results}. This highlights the superior generalization capability of our method compared to existing methods. 
Particularly noteworthy is the significant increase in mIoU for unseen classes in the VOC dataset \textbf{9.6\%} and for unseen classes in the COCO dataset \textbf{2.4\%}

Fig.~\ref{fig:vis} showcases the segmentation outcomes of the ZegCLIP~\cite{zegclip} and our proposed SPT-SEG, both on seen and unseen classes. With the integration of our proposed designs, SPT-SEG demonstrates impressive segmentation capabilities on both seen and unseen classes, effectively distinguishing similar unseen classes. 
For example, our approach effectively segments small target 'sport ball' objects and achieves full recognition of the unseen class 'playing field' (Fig.~\ref{fig:vis}(1)). Furthermore, our method successfully discriminates  ``plastic'' classes from skateboard regions (Fig.~\ref{fig:vis}(2)), and accurately segments ``dog'' instances bearing resemblance to ``horses'' (Fig.~\ref{fig:vis}(3)).
Overall, SPT-SEG completely segments the unseen classes(``playing field'', ``plastic'') and significantly outperforms other methods in terms of segmentation details. These results confirm the effectiveness of our proposed method in achieving superior segmentation performance, especially for unseen classes.

\subsection{Ablation Study}\label{subsec: ablation}
\subsubsection{Detailed results of applying designs on baseline}

To demonstrate the effectiveness of our proposed designs, we further report the improvements of applying designs on baseline (ZegCLIP) in Tab.~\ref{tab:designs}. 
The addition of the SPT significantly enhances the model's performance on unseen data. When both SPT and SGD are utilized, the SPT-SEG model exhibits excellent results on the VOC test dataset. 

\begin{table}
\centering
\begin{tabular}{cll|ccc} 
\hline
\multirow{2}{*}{\textbf{Bas.}} & \multicolumn{1}{c}{\multirow{2}{*}{\textbf{SPT}}} & \multicolumn{1}{c|}{\multirow{2}{*}{\textbf{SGD}}} & \multicolumn{3}{c}{\textbf{PASCAL VOC 2012}}         \\ 
\cline{4-6}
                               & \multicolumn{1}{c}{}                              & \multicolumn{1}{c|}{}                              & \textbf{mIoU(S)} & \textbf{mIoU(U)} & \textbf{hIoU}  \\ 
\cline{4-6}
\CheckmarkBold              &                                                   &                                                    & 91.9             & 77.8             & 84.3           \\ 
\hdashline
\CheckmarkBold                &  \CheckmarkBold                              &                                                    & 92.6             & 86.7             & 89.6           \\
\CheckmarkBold                    &                                                   & \CheckmarkBold                              & 92.0             & 79.9             & 85.5           \\
\CheckmarkBold                 & \CheckmarkBold                                    &\CheckmarkBold                                   & \textbf{92.9}    & \textbf{87.4}    & \textbf{90.1}  \\
\hline
\end{tabular}
\caption{Quantitative results on VOC dataset to demonstrate the effectiveness of our proposed two designs. Here \CheckmarkBold means that this component is applied. Note that our baseline (Bas.) method is ZegCLIP~\cite{zegclip}. The best results are highlighted in bold.}
\label{tab:designs}
\end{table}

\subsubsection{Effect of the depth of SPT}
Tab.~\ref{tab: depth} demonstrates the impact of SPT insertion positions and layers on SPT-SEG performance. The performance of SPT is notably superior when inserted in the earlier layers compared to the later ones. However, its overall performance is comparable when applied across all layers as well as with its application limited to the first two layers. This finding indicates the greater significance of early transformer layer spectral prompts over later layers' prompts.

\begin{table}
\centering
\begin{tabular}{c|cll} 
\toprule
\multirow{2}{*}{\textbf{Depth}} & \multicolumn{3}{c}{\textbf{PASCAL VOC 2012}}                             \\ 
\cline{2-4}
                                & \multicolumn{1}{l}{\textbf{mIoU(S)}} & \textbf{mIoU(U)} & \textbf{hIoU}  \\ 
\hline
1-6                             & 92.5                                 & 86.4             & 89.3           \\
6-12                            & 92.1                                 & 80.9             & 86.1           \\
1-12                            & 92.6                                 & 86.5             & 89.4           \\
11-12                           & 92.0                                 & 78.3             & 84.6           \\
1-2                             & \textbf{92.9}                        & \textbf{87.4}    & \textbf{90.1}  \\
\hline
\end{tabular}
\caption{{Ablation on Spectral Prompt Tuning depth. The 1-st layer refers to the one closest to input. ViT-B has 12 layers in total.The best results are highlighted in bold.}}
\label{tab: depth}
\end{table}

\subsubsection{Effect of Spectral Guided Decode layers}
To investigate the impact of decoder layers on the performance of SPT-SEG, we conducted an ablation study on decoder layer depth. Tab.~\ref{tab: decode_layer} demonstrates that within our research settings, the model achieved its optimal performance with 3 decoder layers. At this layer depth, the model exhibited excellent performance both at the pixel-level and class-level. However, when the decoder layers were increased to 5, we observed signs of overfitting, resulting in a decline in performance on the test set. Conversely, employing only 1 decoder layer significantly reduced the model's performance.

\begin{table}
\centering
\begin{tabular}{c|cll} 
\toprule
\multirow{2}{*}{\textbf{Layers}} & \multicolumn{3}{c}{\textbf{PASCAL VOC 2012}}                             \\ 
\cline{2-4}
                                 & \multicolumn{1}{l}{\textbf{mIoU(S)}} & \textbf{mIoU(U)} & \textbf{hIoU}  \\ 
\hline
1                                & 91.9                                 & 82.8             & 87.1           \\
3                                & \textbf{92.9}                        & \textbf{87.4}    & \textbf{90.1}  \\
5                                & 92.2                                 & 83.7             & 87.7           \\
\hline
\end{tabular}
\caption{{Ablation on layers of Spectral Guided Decode Layer. The best results are highlighted in bold.}}
\label{tab: decode_layer}
\end{table}

\section{Limitations}
Limited by the recognition capability and resolution of CLIP, pixel classification may be prone to errors in complex scenes such as object occlusion and glass  reflection  (e.g. (Fig.~\ref{fig:vis}(5))). Additionally, the ability to recognize details, such as object edges, also needs improvement. Resolving these limitations and enhancing the robustness of the SPT-SEG method are important directions for future research.

\section{Conclusion}
In this work, we present an efficient one-stage direct zero-shot semantic segmentation method based on the pre-trained vision-language model CLIP. We introduce two innovative designs to transfer image classification capabilities to dense prediction tasks while maintaining a leading edge in zero-shot knowledge. These designs enable us to achieve competitive results on known classes and significantly improve performance on novel classes. To demonstrate the effectiveness of our approach, we comprehensively test its performance on two widely-used benchmark datasets, outperforming the previous state-of-the-art methods.
Our research aims to explore the use of pre-trained visual language models for semantic segmentation. By integrating spectral information and enhancing the capability of CLIP, we successfully apply its zero-shot knowledge to downstream tasks, providing a flexible and accurate solution for zero-shot semantic segmentation.

\section{Acknowledgements}
This work was supported by Beijing Natural Science Foundation No. JQ23014, and in part by the National Natural Science Foundation of China (Nos. $U21A20515$, $62271074$ and $62276031$).

\bibliography{aaai24}

\end{document}